# Robust self-healing prediction model for high dimensional data

Anirudha Rayasam
Samsung Research and Development Institute, Bangalore
aniriudha_r.c@samsung.com

Nagamma Patil
National Institute of Technology Karnataka, Surathkal
nagammapatil@nitk.ac.in

## ABSTRACT
Owing to the advantages of increased accuracy and the potential to detect unseen patterns, provided by data mining techniques they have been widely incorporated for standard classification problems. They have often been used for high precision disease prediction in the medical field, and several hybrid prediction models capable of achieving high accuracies have been proposed. Though this stands true most of the previous models fail to efficiently address the recurring issue of bad data quality which plagues most high dimensional data, and especially proves troublesome in the highly sensitive medical data. This work proposes a robust self-healing (RSH) hybrid prediction model which functions by using the data in its entirety by removing errors and inconsistencies from it rather than discarding any data. Initial processing involves data preparation followed by cleansing or scrubbing through context-dependent attribute correction, which ensures that there is no significant loss of relevant information before the feature selection and prediction phases. An ensemble of heterogeneous classifiers, subjected to local boosting, is utilized to build the prediction model and genetic algorithm based wrapper feature selection technique wrapped on the respective classifiers is employed to select the corresponding optimal set of features, which warrant higher accuracy. The proposed method is compared with some of the existing high performing models and the results are analyzed.

## 1. INTRODUCTION
The wide usage of data mining techniques can be attributed the vast functionalities it offers to perform a varied number of useful tasks. It can effectively help to extract useful knowledge from enormous amount of data which are readily available these days, thus providing worth and also enabling them to promote the efficient usage of other related tasks. Classification and prediction are two types of data analysis that can be used to create classifier models and predict future data trends. Clustering is a process by which data is grouped into clusters according to their similarities or dissimilarities. Feature selection is yet another data mining task used in prediction models to find an optimal set of features from the given set of features. Data cleaning is the necessary precursor of knowledge discovery and data warehouse building. As data collected from a varied number of sources can have issues of missing, erroneous, duplicated or structurally and semantically heterogeneous data etc. data cleansing plays a crucial role in todays world in order for the accurate performance of other complementary mining techniques. Using the combination of these basic techniques powerful hybrid models capable of extensive applications can be developed.

High dimensional datasets like medical datasets, biological sequences etc. have a large set of features and data mining techniques prove particularly very helpful in their analyses and for disease prediction. One such application is prediction of Type-2 diabetes, which is a disease caused due to insulin deficiency and if left unaddressed could prove fatal. Previously several high accuracy systems which employ a varied degree of data mining techniques have been proposed for the early prediction of diabetes on the Pima Indians diabetes data. Most of the prevailing solutions do not strongly address the bad quality of the data and just discard the tuples which seem incomplete or damaged. By doing so a large quantity of information is lost and not considered in building the prediction model, which affects the robustness of the system. In the sense that when instances similar to the excluded tuples occur at a later point in time, then the model would fail to correctly classify them as the model has not been allowed to learn any relevant features to classify instances of this type. Also the ignored tuples are not included during the evaluation of the system, which results in the accuracies showcased tending to be superfluous. Also most times these missing data may greatly influence the feature selection process resulting in crucial attributes being unconsidered. Therefore, it is imperative that we develop mechanisms to effectively incorporate the entire or most of the data in the development of the prediction models. And to ensure that this inclusion does not adversely affect the system, measures to improve data quality are to be taken.

In this work a robust self-healing (RSH) hybrid prediction model is proposed, which is comprehensive and provides improved accuracies for disease prediction. The initial preprocessing of the data is performed consisting of data normalization and grouping, followed by attribute correction. The crux of the prediction model comprises of an ensemble of heterogeneous classifiers, each of which are trained with a specific set of optimal features that are chosen via genetic algorithm based wrapper feature technique. The model is evaluated on the Type-2 diabetes Pima Indians dataset and a comparative analysis is performed with the existing meth-ods.

The organization of the rest of the paper is as follows: an overview of the related work is briefed in Section 2. Section 3 details the work on the proposed model, while Section 4 provides a comparative analysis of the results obtained with the previous models. The conclusion and the future scope of the work is stated in Section 5.

## 2. RELATED WORK

Data mining techniques have been incorporated in the medical domain for the prediction of diseases for a long time and various models have been proposed over time. Initially, several traditional classification techniques were used for the purpose, yielding average accuracies. The successive models resorted to clustering [8][3][21][14] and feature selection [20][6][16] techniques to enhance the achievable accuracies of the classifier models as seen in. The advantages of ensemble idea in supervised learning has encouraged their usage for a long time and boosting has been widely used to improve the accuracy of ensemble models. The following works provides an overview of several ensemble models and their applications [17][19].

Recent work by researchers have shown that the hybrid models have been very prosperous in enhancing the accuracies in disease prediction. By the use of several data mining techniques in collaboration with each other it has been possible to amplify the efficiency of the systems. A hybrid model that uses a multi-objective local search to perfectly balance between local and genetic searches has been described in the work of Ishibuchi et al. [11]. The model proposed by Vafaie et al. in [1] uses genetic search techniques in comparison to greedy search and skilfully gleans an initially unknown search space to bias the successive search into promising subspaces. Several fuzzy hybrid models have also been proposed and have proven to perform well. It can be seen in works of Carlos et al. [15] which uses a fuzzy-genetic approach adopting an evolutionary model to enable classification; Fan et al. [7] that combines soft computing techniques with decision tree tools to diagonise and classify breast cancer and liver disorder; and the work of Amit et al. [4] discussing a fuzzy system developed by heuristically learning from neural networks. The model proposed in [12] also incorporates a hybrid neural network of Artificial Neural Network(ANN) and Fuzzy Neural Network.

More recent work on hybrid models by B.M Patil et al. [9] uses a clustering algorithm as a preprocessing step before the classification process to eliminate the incorrectly clustered tuples. The cleansed data is used to build the classifier model which is then tested on the same cleansed data by k-fold cross validation. The proposed technique results in high accuracies for disease prediction. The model proposed in [18][2] extends the the previous model by B.M Patil et al. through the usage of more efficient clustering techniques for the elimination of outliers and genetic algorithm based wrapper feature selection to select an optimal subset of attributes from the dataset, resulting in further increase of prediction accuracy. Though these models achieve high accuracies they are attained at the cost of robustness of the system and are not reliable. As an alternative to removing tuples completely, data quality can be made better through data scrubbing and attribute correction techniques. Overview and details of the types and classification of data quality issues, various design suggestion, model frameworks and common techniques including the approaches of clustering and association rule mining for data cleansing and attribute correction is detailed in [5][24][23]. A fuzzy data mining technique to mine association rules from quantitative, which we have adopted in this paper is detailed in [10].

## 3. PROPOSED ROBUST SELF-HEALING HYBRID PREDICTION MODEL

In this model the dataset is cleansed in the preprocessing phase which involves attribute correction to logically predict the missing values and fix any inconsistencies or spurious tuples, followed by data normalization which prepares the data to be modeled by the classification phase. An ensemble model consisting of heterogeneous base classifiers is used for the prediction. To further enhance the prediction accuracy boosting is performed individually on each classifiers with varying levels of iteration and then the resultant outputs are combined through maximum vote. Each of the base classifiers utilizes only a fraction of the attributes for modeling chosen through a feature selection process. An overview of the system architecture is provided in Figure 1. Working of each of the components is detailed in the following sections.

### 3.1 Data preprocessing

The raw dataset obtained can have data quality issues on instance level or on the record/schema level. Data mis-entry, redundancy, inconsistent aggregation etc. are some of the instance level issues and the issues on the schema level include uniqueness, referential integrity and naming or structural conflicts. In order to handle these, context-dependent attribute correction is incorporated. Context-dependent implies that the correction is achieved not only with reference to the data values it is similar to, but also depends on the values of the other attributes within the native record unlike context-independent correction in which all record attributes are cleaned in isolation.

#### 3.1.1 Non-numeric attribute correction

The numeric and the non-numeric attributes are handled differently. For the non-numeric attributes all the frequent sets are generated using the apriori algorithm [22]. The association rules are generated for these stets and may have either 1, 2 or 3 predecessors and a single successor. These rules form the set of validation rules. For each of the tuples having attribute values which vary from a validation rule, a check is performed with all successors of the rule. If the resulting normalized Levenshtein distance with any of them is lower than a particular distance threshold then the attribute value of the corresponding tuple is altered. The normalized Levenshtein distance between two strings $a$ and $b$ is calculated as below:

$$NormLev(a,b) = \frac{1}{2} * \left( \frac{LevDist(a,b)}{|a|} + \frac{LevDist(a,b)}{|b|} \right) \quad (1)$$

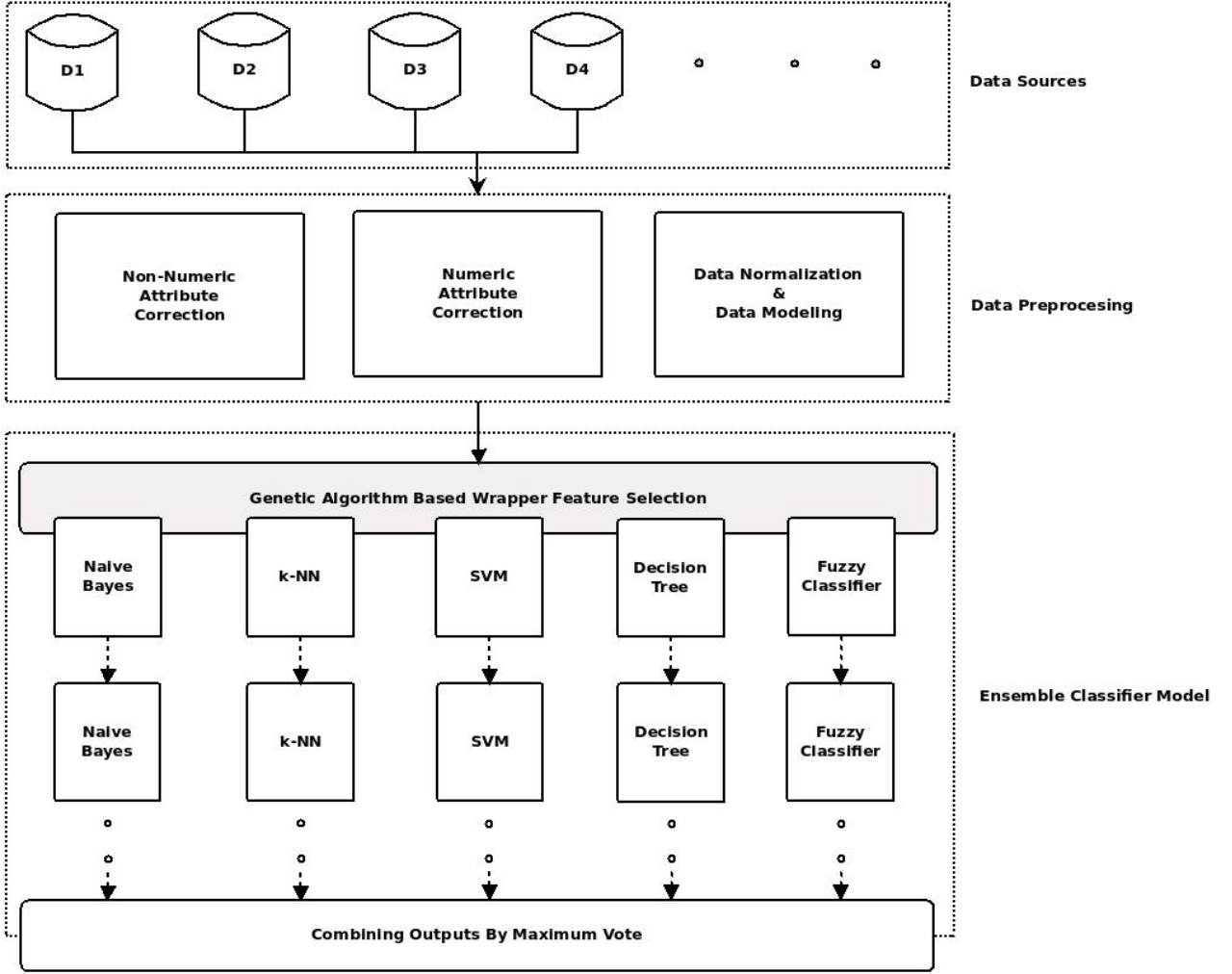

Figure 1: **Overview of the proposed system architecture.**

where *LevDist(a,b)* is the Levenshtein distance between $a$ and $b$; $|a|$ and $|b|$ are the respective attribute lengths.

### 3.1.2 Numeric attribute correction

Fuzzy concepts are used in the apriori algorithm to discover important association rules for quantitative data. All the numeric values for every tuple are transformed into fuzzy sets as described in [10], and the cardinality of each attribute is calculated as the summation of belongingness to each fuzzy set. The attributes are represented by the fuzzy regions that they belong to along with the respective membership values. Cardinality of each fuzzy region is compared to the support threshold value $\alpha$, and if found to be greater then it is included into the *large 1-itemset*. By using these as the seeds, the initial candidate sets are generated and iteratively expanded in a way similar to the apriori algorithm to obtain the *r+1-itemsets*. During each iteration the scalar cardinality of the fuzzy regions is updated and if found higher than $\alpha$ added to the next higher candidate set and it continues until the cardinalities cease to exceed the threshold. From the obtained *q-itemsets* form all the possible combinations of association rules and only the ones with confidence values greater than the confidence threshold $\lambda$ serve as the set of fuzzy association rules. For maximum efficiency the $\alpha$ and $\lambda$ values are set at 2.5 and 0.7 respectively. Similar to the non-numeric attributes all the tuples containing suspicious attribute values are compared to the fuzzy rules that are similar to it and the value is changed to the most probable value.

### 3.1.3 Data normalization

The tuples which continue to have a large number of missing values or errors post attribute correction are removed and the data is normalized. Data normalization is a method of standardization of the values in the dataset to a particular range in order to generalize the values during the clustering and classification processes. It is performed by using z-score normalization as given in the below equation.

$$v^{'} = \frac{v - \mu}{\sigma} \qquad (2)$$

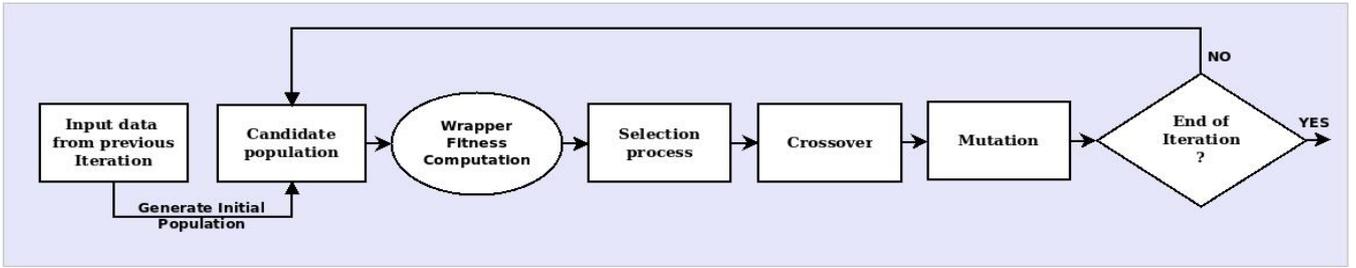

Figure 2: **GA based wrapper feature selection algorithm.**

where $v'$ is the normalized value, $v$ is the experimental value, $\mu$ is the mean and $\sigma$ is the standard deviation.

## 3.2 Feature selection

Genetic algorithm (GA) based wrapper feature selection is adopted for the selection of an optimal feature set. It is applied for each of the base classifiers to ensure a more efficient and accurate prediction model. Figure 2 gives a graphical overview of the GA based wrapper feature selection process. GA works analogous to the evolution process, thus obtaining an incrementally improving solution for a particular problem and in this case as we use a wrapper based approach it strives towards achieving higher accuracies for the chosen classifier. Each solution is regarded as an individual and a set of solutions represents a population of individuals. The best individuals, measured by their fitness, are allowed to evolve within the given population. Fitness in this particular case of feature selection is the ability of an individual (tuple/record) to accurately represent the greater population and thus contribute to higher prediction accuracies. The individuals are modeled as a binary sequence with the length denoting the total number of attributes in the dataset and each bit representing an attribute. A value of 1 or 0 at a particular position indicates the presence or absence respectively for the corresponding attribute. Maximum population size is fixed at 100 and an initial population set is randomly generated, with the successive generations being generated from this by means of crossover, with a probability of 0.75, and mutation, with a probability of 0.03. This final reduced feature set is used to build the base classifier.

## 3.3 Ensemble classifier model

In ensemble methodology several individual classifiers are weighed and combined in order to obtain a single high performing classifier which outperforms the individual classifiers. Single classifiers usually suffer from lack of sufficient training as it would we difficult for a single model to learn all the features of the dataset and having an ensemble of classifiers alleviates this problem by effectively distributing the training task among several classifiers. Using an ensemble further enables larger and more diverse training sets to be learnt, and it also eliminates the problem of local optima and noise or outliers by averaging the predictions. Hence, the incorporation of ensembles in our approach is very advantageous as it can efficiently handle the predictions even in the presence of any remnant errors or outliers in the cleansed dataset.

The main components of a typical ensemble model consists of the training set, base classifiers, which are the building blocks of the ensemble, and the combiner which combines the classifications of the various base classifiers. There is also a virtual component called diversity generator which is responsible for generating diverse base classifiers, as it very necessary for the effective functioning of the ensemble. In the proposed model a combination of heterogeneous classifiers namely Naive Bayes, k-NN, SVM, Decision Trees and Fuzzy classifier, are used as the base classifiers. GA based feature selection described in the previous section is wrapper around each of the individual base classifiers and the dataset is passed through this prior to being sent to the classifiers. A unique optimal feature set is chosen by the feature selector for every base classifier based on which they are trained. This induce more diversity and also fine-tunes the learning of each base classifier model.

One of the main approaches for ensemble based learning which enables higher accuracies to be achievable is model-guided instance selection. In this technique the classifiers constructed in the former iterations assist in the manipulation of the training set for the successive iterations. The latter models are biased to learn the misclassified instances more than the previously correctly classified instances. Boosting, also known as arcing (adaptive re-sampling and combining), is the most well know model-guided instance selection approach that constructs a composite classifier that performs well on the data by iteratively improving the classification accuracy. Local boosting which is based on the popular AdaBoost algorithm is incorporated and it works by giving more focus to the patterns that are harder to classify while ignoring noise and the amount of focus is quantified by a weight assigned to the tuples in the dataset, which determines the probability with which they are picked in the next iterations. Initially all instances are assigned the same weights and after each iteration the misclassified instances are compared with similar instances in the training set to check for their classification status. If these similar instances are also misclassified then the weight for that instance is increased, but if the misclassified instance's neighbors are correctly classified then it is likely that the particular instance is a noisy one and cannot contribute to the learning and thus its weight is reduced. Additionally, the weights for all the correctly classified instances are reduced. The generated classifiers are assigned a rank proportional to the number of correctly classified records and this is utilized for combing the prediction results logically. The generated series of classifiers complement one another and aid in accomplishing better overall results. In this model boosting is performed locally in isolation on each of the base classifiers to different levels of iterations and the results obtained from the various vertical stacks of classifiers are then combined using the majority voting scheme.

Table 1: **Information of modeling metrics and accuracy of base classifiers and previous models**

| Sl.No. | Prediction Model | Total instances | No. of instances retained | No. of features selected | Accuracy |
|---|---|---|---|---|---|
| 1 | Naive Bayes | 726 | 726 | 9 | 76.40% |
| 2 | k-NN | | | | 71.03% |
| 3 | Decision tree | | | | 75.88% |
| 4 | SVM | | | | 76.93% |
| 5 | GWHPM-SVM | 726 | 625 | 5 | 97.86% |
| 6 | SCHPM | 726 | 514 | 5 | 99.37% |
| 7 | HPM | 768 | 433 | 9 | 92.35% |

Table 2: **Information of modeling metrics of each base classifier of the proposed RSH model.**

| Sl.No. | Base Classifier | No. of instances | Optimal subset size | Boosting level | Accuracy | Specificity | Sensitivity | Kappa statistics |
|---|---|---|---|---|---|---|---|---|
| 1 | Naive Bayes | 731 | 6 | 4 | 91.27% | 90.06% | 91.81% | 0.953 |
| 2 | k-NN | | 5 | 3 | | | | |
| 3 | Decision tree | | 4 | 5 | | | | |
| 4 | SVM | | 5 | 3 | | | | |
| 5 | Fuzzy Classifier | | 5 | 3 | | | | |

## 4. RESULTS AND EVALUATION:

Evaluation of the proposed RSH model is performed on the Pima Indians diabetes dataset from the UCI repository [13]. This dataset consists of 726 instances each having 9 attributes qualifying them and 2 classes to either of which they can belong. The performance measures being recorded are *accuracy*, *specificity* and *sensitivity* which are calculated as follows:

$$Accuracy = \frac{TP + TN}{TP + TN + FP + FN} \quad (3)$$

Similarly, specificity and sensitivity are also measured as follows:

$$Specificity = \frac{TP}{TP + FN} \quad (4)$$

$$Sensitivity = \frac{TN}{TN + FP} \quad (5)$$

with *TP*, *FP*, *TN*, *FN* being the count of True Positive, False Positive, True Negative and False Negative values respectively.

Additionally, the *kappa statistics* which measures the agreement between expected chance and the real world value is also considered and is calculated as below:

$$Kappa\ Statistics = \frac{P(A) - P(E)}{1 - P(E)} \quad (6)$$

$$P(A) = \frac{TP + TN}{N} \quad (7)$$

$$P(E) = \frac{(TP + FN) * (TP + FP) * (TN + FN)}{N^2} \quad (8)$$

where *P(A)* is percentage of agreement between the classifier and the true value, *P(E)* is the chance agreement and *N* is the total number of data-points considered to build the classifier.

The accuracies obtained by each of the individual base classifiers, GWHPM [18], SCHPM [2], and the HPM by B.M Patil et al. [9] along with the details regarding the number of eliminated and used tuples for the respective models is tabulated in Table 1. As it can be observed from the table, the GWHPM model uses only 625 tuples and SCHPM model uses an even lesser number of 514 out of the 726 cleaned tuples in the dataset to build the classifier model, while the HPM model uses just 433 tuples, which is a staggering of just 56% of the total number. The excluded tuples are declared as outliers, but nowhere in their work is it justified as to why they are outliers. They are chosen based on a clustering process which brands the all the relatively smaller clusters as outliers, which need not necessarily be true as they could as well be a set of special cases or have minor errors in them on some crucial attributes due to which they were unable to be correctly clustered. Moreover, the similarity parameter for the clustering model need not be coherent to the classification process. Finally, the data similar to the removed instances are omitted from the evaluation as well which essentially biases the model to exhibit high accuracies, which only hold true for the diminished dataset domain. Due to all these reasons and more it is baseless to remove such a great number of instances as it significantly reduces the size of the dataset, concurrently affecting its diversity and richness which in turn adversely affects the effectiveness of the resulting model.

Table 3: **Comparative study with existing method and the proposed RSH on the diabetes dataset.**

| Method | Accuracy(%) | Reference |
|---|---|---|
| **RSH** | **91.27** | **This study** |
| HPM | 92.35 | B.M. Patil et al |
| Hybrid model | 84.5 | Humar Kahramanli |
| Logdisc | 77.7 | Statlog |
| IncNet | 77.6 | Norbert Jankowski |
| DIPOL92 | 77.6 | Statlog |
| Linear discr.analysis | 77.5 - 77.2 | Statlog; Ster & Dobnikar |
| SMART | 76.8 | Statlog |
| GTO DT $5_CV$ | 76.8 | Bennet and Blue |
| kNN, k = 23, Manh, raw, W | 76.6 ± 3.4 | WD-GM, feature weighting 3CV |
| kNN, k = 1:25, Manh, raw | 76.6 ± 3.4 | WD-GM, most cases k = 23 |
| ASI | 76.6 | Ster & Dobnikar |
| Fisher discr. analysis | 76.5 | Ster & Dobnikar |
| MLP + BP | 76.4 | Ster & Dobnikar |
| MLP + BP | 75.8 ± 6.2 | Zarndt |
| NB | 75.5 - 73.8 | Ster & Dobnikar; Statlog |
| kNN, k = 22, Manh | 75.5 | Karol Grudzinski |
| MML | 75.5 ± 6.3 | Zarndt |
| SNB | 75.4 | Ster & Dobnikar |
| BP | 75.2 | Statlog |
| SSV DT | 75.0 ± 3.6 | WD-GM, SSV BS, node 5CV MC |
| kNN, k = 18, Euclid, raw | 74.8 ± 4.8 | WD-GM |
| CART DT | 74.7 ± 5.4 | Zarndt |
| CART DT | 74.5 | Stalog |
| DB-CART | 74.4 | Shang & Breiman |
| C4.5 $5_CV$ | 72.0 | Bennet and Blue |
| CART | 72.8 | Ster & Dobnikar |
| Kohonen | 2.7 | Statlog |
| kNN | 71.9 | Ster & Dobnikar |
| ID3 | 71.7 ± 6.6 | Zarndt |
| IB3 | 71.7 ± 5.0 | Zarndt |
| IB1 | 70.4 ± 6.2 | Zarndt |
| kNN, k = 1, Euclides, raw | 69.4 ± 4.4 | WD-GM |
| kNN | 67.6 | Statlog |
| C4.5 rules | 67.0 ± 2.9 | Zarndt |
| OCN2 | 65.1 ± 1.1 | Zarndt |

In the proposed model the prepared dataset after initially being subject to attribute correction is cleansed to remove any residual inconsistencies, resulting in the removal of 37 instances which were failed to be mended by the correction process and continue to contain a large number of missing values. The distance threshold for comparing Levenshtein distance is set as 0.2 as it resulted in maximum efficiency. Through the development of a more efficient correction process, it could be aimed to further reduce this number and also enhance the quality of correction, and could be taken as future work. Remainder of the 731 corrected and cleansed tuples are fed to the ensemble classifier. For each of the base classifiers coupled with feature selection an optimal subset of attributes are chosen and are utilized for modeling the corresponding classifier and the respective individual details is provided in Table 2. Different iteration levels of boosting was incorporated using local boosting technique for each type of the base classifier individually and the ones which provided the best results are presented in Table 2 along with the achieved performance details. Boosting can affect each combination of dataset and chosen classifier differently and should be tuned according to needs if they are to be used under different circumstances. The accuracy of 91.27% obtained by this model on the Pima Indians diabetes dataset is compared to the previous models on the same dataset, and as seen in Table 3 it fares comparably well with them along with the assurance of being robust.

## 5. CONCLUSION

This paper describes Robust Self-Healing (RSH) a hybrid prediction model which incorporates an ensemble of heterogeneous classifiers to efficiently tackle the classification problem of high dimensional data. Each of the base classifiers in the system are modeled using an optimal subset of attributes which are chosen by the genetic algorithm based wrapper feature selection technique wrapped around the corresponding base classifier. Further, each of the base classifiers are boosted by incorporating local boosting algorithm to achieve higher accuracies. A preprocessing step is included to cleanse and model the data to obtain normalized data of improved quality which is ready for the ensuing feature selection and classifier building phases. Context-dependent attribute correction through the usage of clustering and association rule mining is utilized for data cleansing and smoothing. This process assures that data is rid of errors, duplications, missing fields and spurious or inconsistent values, and ensures that high quality data capable of building robust classifier models is obtainable without requiring to exclude a significant number of tuples. The proposed model is designed to work well with high dimensional data with inherent data quality issues without being dependent on external references and is evaluated on the Pima Indians diabetes dataset from the UCI repository. Based on the comparative analysis of our study with the existing systems it can be concluded that the described RSH hybrid model efficiently classifies with high accuracies comparable to that of previously reigning systems without having to compromise on the robustness. In the future, similar architectural guidelines could be followed to develop systems with more acutely crafted ensemble models involving more appropriate boosting and better data correction techniques which are compatible with other high dimensional data involving complex attributes and having complicated correlations between them.

# 6. REFERENCES

[1] M. Affenzeller, S. M. Winkler, S. W. 0002, and A. Beham. *Genetic Algorithms and Genetic Programming - Modern Concepts and Practical Applications.* CRC Press, 2009.

[2] R. Anirudha, R. Kannan, and N. Patil. Soft computing based hybrid disease prediction method. In *International Conference on Data Mining and Warehousing-ICDMW*, pages 229 – 237. Elsevier, 2014.

[3] C. Blake and C. Merz. UCI repository of machine learning data sets. http://www.ics.uci.edu/~mlearn/MLRepository.html, 1999.

[4] S. Chattopadhyay, D. K. Pratihar, and S. C. D. Sarkar. A comparative study of fuzzy c-means algorithm and entropy-based fuzzy clustering algorithms. *Computing and Informatics*, 30(4):701–720, 2011.

[5] L. Ciszak. Application of clustering and association methods in data cleaning. In *IMCSIT*, pages 97–103. IEEE, 2008.

[6] M. Dash and H. Liu. Feature selection for classification. *Intelligent Data Analysis*, 1(1):131 – 156, 1997.

[7] C.-Y. Fan, P.-C. Chang, J.-J. Lin, and J.-C. C. Hsieh. A hybrid model combining case-based reasoning and fuzzy decision tree for medical data classification. *Appl. Soft Comput.*, 11(1):632–644, 2011.

[8] J. Han and M. Kamber. *Data mining : concepts and techniques.* Kaufmann, San Francisco [u.a.], 2005.

[9] S. Hirano and S. Tsumoto. Cluster analysis of time-series medical data based on the trajectory representation and multiscale comparison techniques. In *ICDM*, pages 896–901. IEEE Computer Society, 2006.

[10] T.-P. Hong, C.-S. Kuo, and S.-C. Chi. Mining association rules from quantitative data. *Intelligent Data Analysis*, 3(5):363 – 376, 1999.

[11] H. Kahramanli and N. Allahverdi. Design of a hybrid system for the diabetes and heart diseases. *Expert Syst. Appl.*, 35(1-2):82–89, 2008.

[12] H. Kahramanli and N. Allahverdi. Rule extraction from trained adaptive neural networks using artificial immune systems. *Expert Syst. Appl.*, 36(2):1513–1522, 2009.

[13] K. Kayaer and T. Yildirim. Medical diagnosis on Pima Indian diabetes using general regression neural networks, 2003.

[14] B. Krishna and B. Kaliaperumal. Efficient genetic-wrapper algorithm based data mining for feature subset selection in a power quality pattern recognition applicationction in a power quality pattern recognition application. *Int. Arab J. Inf. Technol.*, 8(4):397–405, 2011.

[15] A. Leon-Barranco, C. A. R. Garcia, and R. Zatarain-Cabada. A hybrid fuzzy-genetic algorithm. In D.-S. Huang, K. Li, and G. W. Irwin, editors, *ICIC (1)*, volume 4113 of *Lecture Notes in Computer Science*, pages 500–510. Springer, 2006.

[16] H. Liu and L. Yu. Toward integrating feature selection algorithms for classification and clustering. *IEEE Trans. on Knowl. and Data Eng.*, 17(4):491–502, Apr. 2005.

[17] S. Pobi. *A Study of Machine Learning Performance in the Prediction of Juvenile Diabetes from Clinical Test Results.* University of South Florida, 2006.

[18] A. RC, R. Kannan, and N. Patil. Genetic algorithm based wrapper feature selection on hybrid prediction model for analysis of high dimensional data. In *International Conference on Industrial and Information Systems.* IEEE, 2014.

[19] L. Rokach. Ensemble-based classifiers. *Artif. Intell. Rev.*, 33(1-2):1–39, Feb. 2010.

[20] M. Sun, L. Xiong, H. Sun, and D. Jiang. In *WGEC*.

[21] H. Temurtas, N. Yumusak, and F. Temurtas. A comparative study on diabetes disease diagnosis using neural networks. *Expert Syst. Appl.*, 36(4):8610–8615, 2009.

[22] G. I. Webb. Association rules. In D. N. Ye, editor, *The Handbook of Data Mining, Chapter 2*, pages 25 – 39. Lawrence Erlbaum Associates, 2003.

[23] H. Yan and X. Diao. The design and implementation of data cleaning knowledge modeling. In *Proceedings of the 2008 International Symposium on Knowledge Acquisition and Modeling*, KAM '08, pages 177–179, Washington, DC, USA, 2008. IEEE Computer Society.

[24] H. Yu, Z. Xiao-yi, and Y. Zhen. A universal data cleaning framework based on user model. In *ISECS International Colloquium on Computing, Communication, Control, and Management*, pages 97–103. IEEE, 2009.